\documentclass{article}
\usepackage{ijcai23}

\usepackage{times}
\usepackage{soul}
\usepackage{url}
\usepackage[hidelinks]{hyperref}
\usepackage[utf8]{inputenc}
\usepackage[small]{caption}
\usepackage{graphicx}
\usepackage{amsmath}
\usepackage{amsthm}
\usepackage{booktabs}
\usepackage{algorithm}
\usepackage{algorithmic}
\usepackage[switch]{lineno}

\usepackage{amssymb}
\usepackage{multirow}
\usepackage{multicolrule}
\usepackage{bm}
\usepackage{booktabs}
\usepackage{threeparttable}
\usepackage{threeparttable}

\title{Knowledge Graph Completion based on Tensor Decomposition for Disease Gene Prediction}		

\author{
Xinyan Wang$^1$\and
Ting Jia$^1$\and
Chongyu	Wang$^1$\and
Kuan Xu$^1$\and
Zixin Shu$^1$\and
Jian Yu$^1$\and

Kuo Yang$^{1*}$\And
Xuezhong Zhou$^{1*}$
\affiliations
$^1$Institute of Medical Intelligence, School of Computer and Information Technology, Beijing Jiaotong University, Beijing 100044, China
\emails
\{xinyan\_wang, 19120365\}@bjtu.edu.cn,
wangchongyu2021@163.com
\{xukuan, 18112032, jianyu, yangkuo, xzzhou\}@bjtu.edu.cn,
}

\begin{document}

\maketitle
\begin{abstract}
Accurate identification of disease genes has consistently been one of the keys to decoding a disease's molecular mechanism. Most current approaches focus on constructing biological networks and utilizing machine learning, especially, deep learning to identify disease genes, but ignore the complex relations between entities in the biological knowledge graph. In this paper, we construct a biological knowledge graph centered on diseases and genes, and develop an end-to-end \textbf{K}nowledge graph completion model for \textbf{D}isease \textbf{Gene} Prediction using interactional tensor decomposition (called KDGene). KDGene introduces an interaction module between the embeddings of entities and relations to tensor decomposition, which can effectively enhance the information interaction in biological knowledge. Experimental results show that KDGene significantly outperforms state-of-the-art algorithms. Furthermore, the comprehensive biological analysis of the case of diabetes mellitus confirms KDGene's ability for identifying new and accurate candidate genes. This work proposes a scalable knowledge graph completion framework to identify disease candidate genes, from which the results are promising to provide valuable references for further wet experiments.
\end{abstract}

\section{Introduction}

Unraveling the molecular mechanism of disease is essential for realizing precision medicine \cite{precision_medicine}. One of the main goals is to identify the causing genes of disease. Traditional methods of identifying disease-causing genes (e.g., Genome-wide association study \cite{gwsa}) are mainly obtained through experiments, which are extremely time-consuming and labor-intensive \cite{labor}.

With the completion of the Human Genome Project and the maturity of high-throughput sequencing technology \cite{HGP}, a growing body of computing-based disease gene prediction methods have been developed, which are proven effective \cite{luo2021predicting}. Compared with traditional experiments, computing-based methods can significantly save resources and reduce experimental errors \cite{bc05,bc02}. Typical studies include network propagation methods (\cite{propagation}), clustering or classification methods (\cite{clustering}), network features (\cite{wu2008network,jalilvand2019disease}), and network embedding methods (\cite{network,HerGePred}). Neural network methods have also been applied to disease gene prediction and obtained high performance \cite{PDGNet,Renet,bc04}.

In recent years, Knowledge Graphs (KGs) have been successfully applied to life science research \cite{KGNN}. KG is a semantic network that reveals the relations between entities, which can formally describe things and their relations in the real world. In KGs, nodes represent entities or concepts, and edges are composed of attributes or relations. Knowledge exists in the form of triples \cite{review}. Inferring unknown facts from those already in KG is called KG Completion (KGC). Performing better in existing KGC models, KG Embedding (KGE) based methods learn the latent representation of entities and relations in continuous vector space \cite{survey}. Network Embedding (NE) assigns nodes in a network to low-dimensional representations and preserves the network structure effectively \cite{node2vec}. The main difference between KGE and NE is that the latter focuses on the topology of the network, while KGE focuses on the internal information of different relations and the semantic connotation of facts.

A few studies have explored KGE-based methods individually for disease gene prediction. They tend to adopt existing KGE models from the general domain \cite{DG-KGE2,DG-KGE3} or use external information, such as the textual description of biological entities \cite{DG-KGE1}. Although the conventional KGE models have been proven to be useful for inferring new biological relations, their performance with biological data is not as satisfactory as that of general-domain KGs \cite{DG-KGE4}. One of the key points is how to model KGE in the process of disease gene prediction to accurately capture the interaction between biological entities (such as Protein-Protein Interactions) \cite{bc03,zhu2022multimodal}, so that diseases and genes can be learned with more comprehensive biological features.

To address these issues, we first integrated multiple relations centered on diseases and genes from biomedical knowledge bases to construct a large-scale biological KG, and develop an end-to-end \textbf{K}nowledge graph completion model using an interactional tensor decomposition to identify \textbf{D}isease-\textbf{Gene} associations, called KDGene. KDGene introduces a gating mechanism-based interaction module between the embeddings of entities and relations to tensor decomposition, which can effectively enhance the information interaction in biological knowledge. Perceiving related knowledge, the model is capable of learning the connotation of different relations and endows biological entities and relations with more comprehensive and precise representations, which is beneficial to disease gene prediction. Experimental results show that KDGene performs optimally among existing disease gene prediction methods. In particular, compared with conventional KGE methods, KDGene realizes an average improvement of over 20\% on HR and MAP metrics. Meanwhile, we evaluate the impacts of KGs composed of knowledge with different relation types and degrees of confidence on KDGene's performance. In summary, the main contributions of our work are three-fold:

\begin{enumerate}
	\item We construct a biological knowledge graph centered on diseases and genes, then adopt a scalable end-to-end KGC framework to predict disease genes.
	\item We propose a novel KGC model, called KDGene, specifically for disease gene prediction. The model introduces an interaction module to tensor decomposition, which effectively enhances the information interaction between biological knowledge.
	\item Our KDGene achieves state-of-the-art on disease gene prediction. The biological analysis of diabetes mellitus also confirms KDGene's ability to identify new and accurate candidate genes.

\end{enumerate}

\section{Related Work}

\subsection{Disease Genes Prediction Models}
Researchers have proposed various computing-based Disease Gene Prediction (DGP) methods, mainly divided into four categories: (1) Network Propagation methods. The network propagation models are based on the classic random walk algorithm for the most part, and they are common in early disease gene prediction tasks \cite{PRINCE,DADA}. (2) Methods based on Network Features. These methods usually use the constructed network to obtain the topological feature information of nodes, then calculate the correlation between a query disease and candidate genes, completing the prediction by sorting the gene list \cite{nf}. (3) Supervised Learning methods such as classification \cite{MapGene}.

And (4) Network Embedding and Deep Learning methods. These methods have gained wide attention in recent years \cite{ne,bc01}. HerGePred \cite{HerGePred} is a heterogeneous disease-gene-related network embedding representation framework for disease gene prediction which can realize similarity prediction and a random walk with restart on a reconstructed heterogeneous disease-gene network. GLIM \cite{GLIM} can systematically mine the potential relationships between multilevel elements by embedding the features of the human multilevel network through contrastive learning. With the development of deep learning technology, researchers have tried to build specific neural network models to predict disease genes \cite{PDGNet}.

\subsection{KGE Models}
Recently, growing amounts of work have been proposed to learn distributed representations for entities and relations in KGs, which fall into three major categories \cite{review}: (1) Geometric Models. They utilize distance-based scoring functions to measure the plausibility of facts, which interpret the relation as a geometric transformation in the latent space \cite{Rotate}. The most representative model, TransE \cite{TransE}, regards the relation \textit{r} in each triple \textit{(h, r, t)} as a translation from the head entity \textit{h} to the tail entity \textit{t}, by requiring the tail entity embedding lies close to the sum of the head entity and relation embeddings. (2) Deep Learning Models. These methods learn the embedding vectors of entity and relation using multi-layer neural networks \cite{ConvE}.

And (3) Tensor Decomposition Models. In Canonical Polyadic (CP) decomposition \cite{CPdecomposition}, a tensor can be decomposed into a set of matrices, where each row in the matrix represents an embedding vector of entity or relation. DistMult \cite{DistMult}, a special case of CP decomposition, forces all relation embeddings as diagonal matrices, which reduces the space of parameters and makes the model easier to train. ComplEx \cite{ComplEx} introduces asymmetry into tensor decomposition by adding complex-valued embeddings so that it can simulate asymmetric relations. Since current implementations of CP are lagging behind their competitors, CP-N3 \cite{CP-N3} uses a tensor nuclear p-norm as a regularizer to break through the limitations of CP and obtain good performance.

\subsection{DGP models combined with KGE}
At present, the DGP methods combined with KGE have not been fully exploited. KGED \cite{DG-KGE1} is a convolutional neural network-based KGE model, which is based on a biological KG with entity descriptions to infer relationships between biological entities. Since KGED is used to predict gene-gene relations to generate gene interaction networks for diseases, it is not an end-to-end model for DGP. And it requires textual descriptions of entities, which may introduce noise and are not simple to obtain. In \cite{DG-KGE2}, KGE is adopted to predict disease genes directly. However, this work applies conventional methods in the KGC task without comparison with other DGP models.

\begin{figure*}[h!]
	\centering
	\includegraphics[scale=0.082]{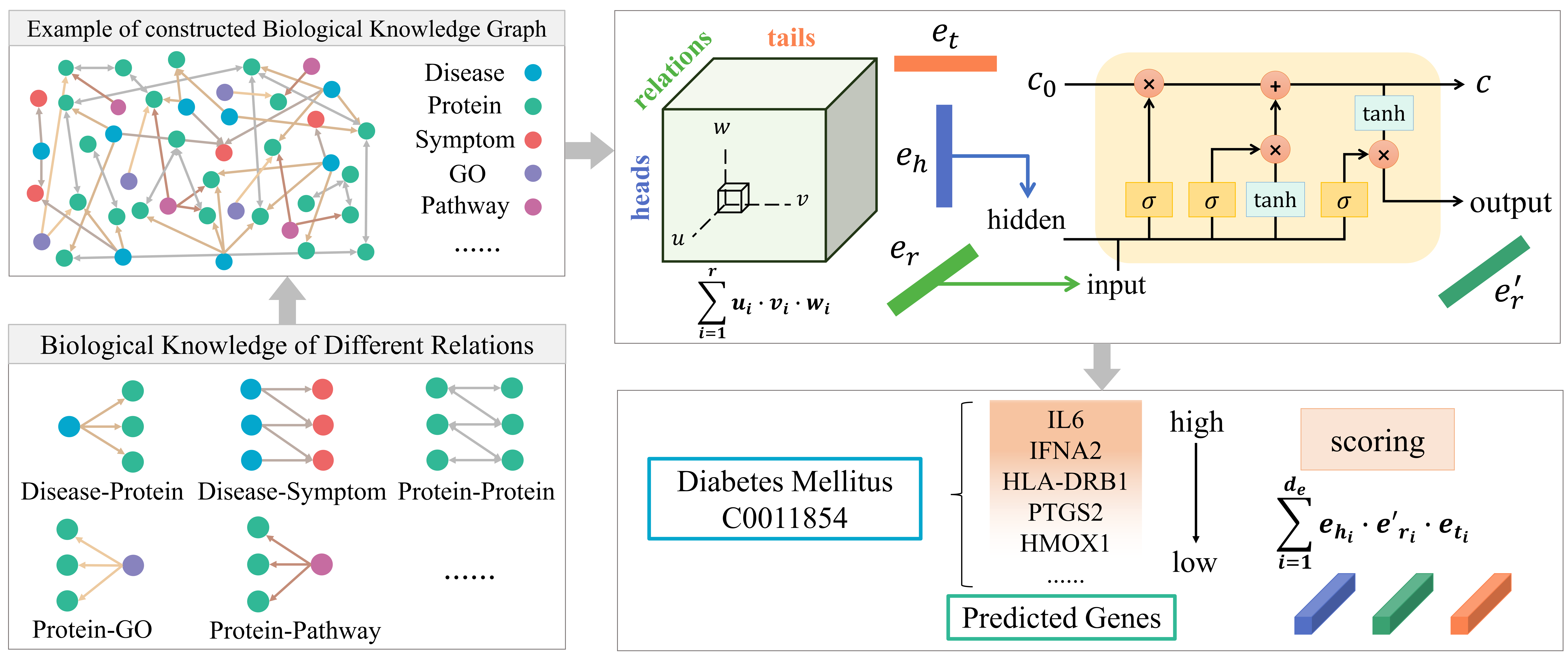}
	\caption{ Visualization of the KDGene architecture. After constructing the biological KG, a triple is represented as $(h, r, t)$, with two entities $h, t$ and a relation $r$. We use $\bm{e_h}, \bm{e_t} \in \mathbb{R}^{d_e}$ to denote the embeddings of head and tail entities and $\bm{e_r} \in \mathbb{R}^{d_r}$ to represent the relation embeddings. When the embeddings $\bm{e_h}, \bm{e_r}, \bm{e_t}$ are trained, taking the relation embedding $\bm{e_r}$ as the input, the head entity embedding $\bm{e_h}$ as the hidden layer, we use the interaction module to obtain the updated relation embedding $\bm{e_r'} \in \mathbb{R}^{d_e}$. Then the scoring function of triple $(h, r, t)$ is calculated by $\bm{e_h}$, $\bm{e_r'}$ and $\bm{e_t}$. After training, for a query disease, score all candidate genes and rank by descending as the prediction results.}
	\label{fig:overview}
\end{figure*}

\section{Preliminaries}

\textbf{\textit{Knowledge Graph $\mathcal{G}$.}}  A knowledge graph can be denoted as $\mathcal{G} = \{ \mathcal{E}, \mathcal{R}, \mathcal{T} \}$ where $\mathcal{E}$ and $\mathcal{R}$ are the entity set and relation set, respectively. And $ \mathcal{T} = \{ (h, r, t) \in \mathcal{E} \times \mathcal{R} \times \mathcal{E} \}$ denotes the triple set which consists of all the triple facts in $\mathcal{G}$. When constructing a biomedical KG, we integrate knowledge from different biological databases in the form of triples and add them to KG. Correspondingly, the associated entity set and relation set are generated.

\noindent \textbf{\textit{Knowledge Graph Completion.}} The task of KGC, also known as Link Prediction, is to either predict unseen relations \textit{r} between two existing entities: (\textit{h}, ?, \textit{t}), or predict entities when the triple's relation and another entity are given: (\textit{h}, \textit{r}, ?) or (?, \textit{r}, \textit{t}). For Disease Gene Prediction, since triples of this kind of facts are in the form (disease, disease\_gene, gene), we focus on the second mode to predict the tail entity (gene) given the head entity (disease) and relation (disease\_gene). 

In this paper, we adopt the improved tensor decomposition-based model under the framework of KGC, in which the triple (\textit{h}, \textit{r}, \textit{t}) can be represented as an element of the binary third-order entity-relation tensor $\mathcal{X}^{\textit{N} \times \textit{M} \times \textit{N}}$, where $\textit{N}=|\mathcal{E}|$ is the total number of entities and $\textit{M}=|\mathcal{R}|$ the number of relations. In the entity-relation tensor $\mathcal{X}$, $\mathcal{X}_{ikj}$ denotes that there is a \textit{k}-th relation between the \textit{i}-th entity and the \textit{j}-th one, which is:

\begin{equation}
\begin{aligned}
\mathcal{X}_{i k j}= \begin{cases}1, & \text { if }\left(h_{i}, r_{k}, t_{j}\right) \in \mathcal{G} \\ 0, & \text { if }\left(h_{i}, r_{k}, t_{j}\right) \notin \mathcal{G}\end{cases}
\end{aligned}
\end{equation}

Therefore, tensor decomposition-based algorithms can infer a predicted tensor $\mathcal{\widehat{X}}$ that approximates $\mathcal{X}$. To predict the candidate genes of a disease, queries like (\textit{i}, \textit{k}, ?) are answered by ordering gene entities $j'$ by decreasing scoring values of $\widehat{\mathcal{X}}_{i k j'}$. Note that we propose a scalable KGC framework for disease gene prediction which means the KGE model can be replaced by others.

\section{Methodology}
We present KDGene, an end-to-end knowledge graph completion model based on interactional tensor decomposition, 

\begin{table}[!ht]
\small
    \centering
    \begin{tabular}{lc|lc}
    \toprule
        Entity Type & quantity & Relation Type & quantity  \\
    \midrule
   		Disease & 22,697  & Disease-Protein & 117,738 \\
   		Protein & 21,616 & Protein-Protein & 841,068 \\
   		Symptom & 2,504 & Disease-Symptom & 184,831 \\
   		GO & 1,207 & GO-Protein & 61,634 \\
   		Pathway & 316 & Pathway-Protein & 25,813 \\
		\midrule
		  Total	& 48,340 & Total & 1,231,084 \\
    \bottomrule
    \end{tabular}
    \caption{The scale of our constructed biological knowledge graph related to diseases and genes.}
    \label{tab:KGscale}
\end{table}

\noindent and formulate the problem of disease gene prediction as an LP task in KG. Figure \ref{fig:overview} shows an overview of KDGene, and the whole framework consists of the following three parts:

\begin{enumerate}
    \item Biological KG Construction. We collected and integrated multiplex relations, e.g., disease-gene, disease-symptom, protein-protein, protein-GO, and protein-pathway from well-known biomedical databases and construct a biological KG centered on diseases and genes.
    \item Representation learning of entities and relations in the biological KG. We do not directly utilize the existing KGE methods but improve the tensor decomposition model by introducing an interaction module, which can enhance the information interaction in biological knowledge and learn the more dedicated representation of relations.
    \item Scoring and ranking of disease candidate genes. Based on the representation features learned in the second step, we score and sort all the candidate genes of a disease according to Equation \ref{score}, and obtain predicted results of candidate genes.
\end{enumerate}

\subsection{Biological KG Construction}
To learn more comprehensive representations of diseases and genes, we introduce the knowledge of different relation types to construct a biological KG. Regarding diseases, the disease-symptom relations from SymMap \cite{SymMap} are introduced into the KG. Regarding genes, we introduce the Protein-Protein Interactions (PPI) from STRING \cite{STRING}, the Protein-GO relations from \cite{GO}, and the Protein-Pathway relations from KEGG\cite{KEGG}. Table \ref{tab:KGscale} shows the scale of the biological KG.

In our framework, there are no restrictions on the entity type and relation type which means the construction of the KG is flexible. When others use it, the disease-gene relation facts can be added or subtracted from the KG to complete the training according to the demand. As the amount of knowledge in biomedical databases grows, abundant facts about new types of relations can be continuously added to the KG.

\subsection{CP-N3}
The KGC task can be regarded as a 3D binary tensor completion problem, where each slice is the adjacency matrix of one relation type in the KG. It is a natural solution to apply tensor decomposition to the KGC task, which is simple, expressive, and can achieve state-of-the-art results in general-domain KGs. Here, we take the typical CP-N3 model as an example and further introduce the interaction module on this basis to predict disease-gene associations.

\textbf{CP-N3} \cite{CP-N3} is based on CP decomposition \cite{CPdecomposition}, which decomposes a high-order tensor $\mathcal{X} \in \mathbb{R}^{n_1 \times n_2 \times n_3}$ into several $r$ rank one tensors $u_i \in \mathbb{R}^{n_1}, v_i \in \mathbb{R}^{n_2}, w_i \in \mathbb{R}^{n_3}$ ($\otimes$ denotes the tensor product):

\begin{equation}
\begin{aligned}
\mathcal{X} \approx \sum_{i=1}^r u_i \otimes v_i \otimes w_i.
\end{aligned}
\end{equation}

\subsection{Interaction Module}
Introducing the interaction module aims to equip KGE models, TD-based methods in particular, with better biomedical knowledge perception. That is, the model should learn more precise representations of entities and relations. To deal with the problem of long-term dependencies, Hochreiter and Schmidhuber proposed long short-term memory (LSTM) \cite{LSTM}. They improved the remembering capacity of the standard recurrent cell by introducing a “gate” into the cell in which the gate mechanism can choose which information enters the next cell \cite{LSTMreview}. We adopt the vanilla LSTM cell \cite{LSTM2} consisting of an input gate, an output gate, and a forget gate. The activation process of LSTM is as follows:

First, the forget gate $f$ and the input gate $i$ at the time step $t$ are computed by

\begin{equation}
\begin{aligned}
f_t & =\sigma\left(W_{f h} h_{t-1}+W_{f x} x_t+b_f\right), \\
i_t & =\sigma\left(W_{i h} h_{t-1}+W_{i x} x_t+b_i\right), \\
\tilde{c}_t & =\tanh \left(W_{\tilde{c} h} h_{t-1}+W_{\tilde{c} x} x_t+b_{\tilde{c}}\right), \\
\end{aligned}
\end{equation}

where $\sigma$ is the logistic sigmoid function, and $x_t$ is the current input. For the forget gate, the LSTM unit determines which information should be removed from its previous cell states $h_{t-1}$. The candidate memory cell is also added to the cell state through a TanH Layer. All the $W$ are weights that need to be learned, while $b$ represents the bias vector associated with this component. Then, the cell state is updated by 

\begin{equation}
\begin{aligned}
c_t & =f_t \circ c_{t-1}+i_t \circ \tilde{c}_t, \\
o_t & =\sigma\left(W_{o h} h_{t-1}+W_{o x} x_t+b_o\right), \\
h_t & =o_t \circ \tanh \left(c_t\right).
\end{aligned}
\end{equation}

$o_t, h_t$ are the outputs at the current time, and $\circ$ is the Hadamard product. In this intuitionistic structure, the control of the forget gate can save the previous information, and the control of the input gate can prevent the current irrelevant information from being added to the cell. The information in each part sufficiently interacts with others, so we utilize this simple and effective structure as our interaction module.

\subsection{KDGene}
We present KDGene, a knowledge graph completion model that introduces the interaction module into CP-N3, which applies to disease gene prediction. In the following, a triple is represented as $(h, r, t)$, with two entities $h, t \in E$ (the set of entities) and a relation $r \in R$ (the set of relations). We use $\bm{e_h}, \bm{e_t} \in \mathbb{R}^{d_e}$ to denote the embeddings of head and tail entities and $\bm{e_r} \in \mathbb{R}^{d_r}$ to represent the relation embeddings.

Instead of adopting the translation-based principle $\bm{h} + \bm{r} = \bm{t}$ in TransE \cite{TransE}, we use the gating mechanism as the entity-to-relation translation. When the embeddings $\bm{e_h}, \bm{e_r}, \bm{e_t}$ are trained, taking the relation embedding $\bm{e_r}$ as the input, and the head entity embedding $\bm{e_h}$ as the hidden layer, we use an LSTM cell to obtain the updated relation embedding $\bm{e_r'} \in \mathbb{R}^{d_e}$. The calculation process is as follows:

\begin{equation}
\begin{aligned}
f & =\sigma\left(W_{f h} \bm{e_h}+W_{f x} \bm{e_r}+b_f\right), \\
i & =\sigma\left(W_{i h} \bm{e_h}+W_{i x} \bm{e_r}+b_i\right), \\
\tilde{c} & =\tanh \left(W_{\tilde{c} h} \bm{e_h}+W_{\tilde{c} x} \bm{e_r}+b_{\tilde{c}}\right), \\
c & =f \circ c_0 + i \circ \tilde{c}, \\
\bm{e_r'} & =\sigma\left(W_{o h} \bm{e_h}+W_{o x} \bm{e_r}+b_o\right), \\
\end{aligned}
\end{equation}

where all $W$ are weight matrices and $b$ are bias vectors learned in the training process. The initial input of the cell state is set to 0.

After getting the updated relation embedding $\bm{e_r'}$, we define the scoring function of a triple $(h, r, t)$ for KDGene as:

\begin{equation}
\begin{aligned}
\label{score}
\phi(h, r, t) = \sum_{i=1}^{d_e} e_{hi} \otimes e_{ri}' \otimes e_{ti}.
\end{aligned}
\end{equation}

In CP-N3, the embedding dimensions of entities and relations must be the same, resulting in a lot of parameter redundancy for those datasets with very different numbers of entities and relations. After introducing the interaction module, the dimensions of entities and relations can be different, which significantly improves the operability and flexibility of KDGene. More importantly, through the gating mechanism of LSTM, entities, and relations are learned with more precise representations, which will benefit disease gene prediction.

\subsection{Training and Prediction}
We use the standard data augmentation techniques \cite{CP-N3} of adding reciprocal predicates in the original training set $S$ and get $S'$, i.e. add $(t, r^{-1}, h)$ for every $(h, r, t)$. Besides, we follow the 1-N scoring introduced by \cite{ConvE}, that is, we take one $(h, r)$ pair and score it against all entities $t' \in E$ simultaneously. We train our model with the full multiclass log-loss:

\begin{equation}
\begin{aligned}
\mathcal{L} = \sum_{(h,r,t) \in S'}( - \phi(h, r, t) + log(\sum_{t' \in E}exp(\phi(h, r, t')))).
\end{aligned}
\end{equation}

where $\mathcal{L}$ is the loss function that should be minimized. For KDGene, we follow the N3 regularization used in CP-N3 \cite{CP-N3}, and the loss function for KDGene is as follows:

\begin{align}
\mathcal{L} =  & \sum_{(h,r,t) \in S'}( - \phi(h, r, t) + log(\sum_{t' \in E}exp(\phi(h, r, t'))) \nonumber \\
				  + & \lambda \sum_{i}^{d_e}(|e_{hi}|^{3} + |e_{ri}'|^{3} + |e_{ti}|^{3})).
\end{align}

After training, for disease gene prediction, we take $(h,r)$ pairs, where the head entity is the query disease, and the relation is disease-gene, and then score all candidate genes that are not in the training set. The list of genes with scores from high to low is the prediction result of the candidate genes.

\begin{table*}[!ht]
\small
    \centering

    \begin{tabular}{lcccccccc}
    \toprule
        DGP Models & HR@1 & HR@3 & HR@10 & HR@50 & MAP@1 & MAP@3 & MAP@10 & MAP@50 \\ 
			\midrule
        DADA & 0.012  & 0.025  & 0.047  & 0.107  & 0.045  & 0.044  & 0.049  & 0.053  \\ 
        GUILD & 0.023  & 0.032  & 0.049  & 0.107  & 0.073  & 0.076  & 0.080  & 0.084  \\ 
        RWRH & 0.082  & 0.153  & 0.269  & 0.486  & 0.297  & 0.268  & 0.272  & 0.286  \\ 
        PDGNet & 0.020  & 0.031  & 0.045  & 0.068  & 0.094  & 0.056  & 0.044  & 0.043  \\ 
        PRINCE & 0.006  & 0.011  & 0.024  & 0.074  & 0.025  & 0.026  & 0.028  & 0.031  \\ 
        RWR\_PPI & 0.070  & 0.148  & 0.271  & 0.474  & 0.257  & 0.241  & 0.255  & 0.270  \\ 
        RWR\_HMLN & 0.094  & 0.180  & 0.304  & 0.502  & 0.342  & 0.303  & 0.306  & 0.320  \\ 
        GLIM\_DG & \underline{0.105}  & \underline{0.194}  & \underline{0.312}  & \underline{0.508}  & \underline{0.383}  & \underline{0.335}  & \underline{0.329}  & \underline{0.342}  \\ 
        KDGene (ours) & \textbf{0.126}  & \textbf{0.243}  & \textbf{0.416}  & \textbf{0.620}  & \textbf{0.406}  & \textbf{0.365}  & \textbf{0.361}  & \textbf{0.370}  \\ 
        Improvement & +19.96\% & +25.16\% & +33.25\% & +22.04\% & +5.82\% & +8.91\% & +9.48\% & +8.12\% \\ 
			\bottomrule
    \end{tabular}

    \begin{tabular}{lcccccccc}
    		\toprule
        KGC Models & HR@1 & HR@3 & HR@10 & HR@50 & MAP@1 & MAP@3 & MAP@10 & MAP@50 \\ 
			\midrule
        TransE & 0.086  & 0.160  & 0.272  & 0.472  & 0.278  & 0.243  & 0.241  & 0.252  \\ 
        RotatE & 0.085  & 0.159  & 0.272  & 0.477  & 0.275  & 0.241  & 0.241  & 0.252  \\ 
        DistMult & \underline{0.107}  & \underline{0.200}  & 0.309  & 0.406  & \underline{0.346}  & \underline{0.301}  & \underline{0.292}  & \underline{0.299}  \\ 
        ComplEx & 0.103  & 0.193  & \underline{0.317}  & \underline{0.515}  & 0.331  & 0.288  & 0.281  & 0.291  \\ 
        TuckER & 0.096  & 0.182  & 0.288  & 0.394  & 0.308  & 0.269  & 0.261  & 0.269  \\ 
        CP-N3 & 0.090  & 0.165  & 0.273  & 0.471  & 0.290  & 0.249  & 0.244  & 0.254  \\ 
        KDGene (ours) & \textbf{0.126}  & \textbf{0.243}  & \textbf{0.416}  & \textbf{0.620}  & \textbf{0.406}  & \textbf{0.365}  & \textbf{0.361}  & \textbf{0.370}  \\ 
        Improvement & +17.24\% & +21.44\% & +31.15\% & +20.35\% & +17.23\% & +21.12\% & +23.52\% & +23.59\% \\ 
			\bottomrule
    \end{tabular}

    \caption{Disease gene prediction results on DisGeNet. The table consists of two parts, the upper part is about the baselines for typical Disease Gene Prediction (DGP) models, and the lower is about the baselines of KGC models. \textbf{Bold} numbers are the best results of all and \underline{underline} numbers are the best results of baseline models.  Improved results of the last row compare the performance of KDGene with the best of baseline models.}
    \label{tab:result}
\end{table*}

\section{Experiments}

\subsection{Experimental setting}
\textbf{Dataset.} We select curated disease-gene associations from the DisGeNet database \cite{DisGeNet} as a benchmark dataset and apply the conventional 10-fold cross validation to evaluate the disease gene prediction algorithms. For each fold, there are 117,738 disease-gene associations in the training set and 13,082 in the testing set. 

\noindent \textbf{Baselines.} For baselines, comparisons with existing disease gene prediction algorithms are essential. Typical models including DADA \cite{DADA}, GUILD \cite{GUILD}, RWRH \cite{RWRH}, PDGNet \cite{PDGNet}, PRINCE \cite{PRINCE}, GLIM \cite{GLIM} are our baseline. In addition, since we formulate disease gene prediction as the KGC task, and propose a novel KGC method, KDGene should also be compared with existing KGC models. We experiment with six popular KGE baselines: TransE \cite{TransE}, RotatE \cite{Rotate}, DistMult \cite{DistMult}, ComplEx \cite{ComplEx}, TuckER \cite{TuckER} and CP-N3 \cite{CP-N3}.

\noindent \textbf{Evaluation Metrics.} Following GLIM \cite{GLIM}, we select the hit ratio (HR@N) and mean average precision (MAP@N) as evaluation metrics (where N=1,3,10,50). For both HR and MAP, higher values indicate higher predictive performance. 

\noindent \textbf{Implementation Details.} 
We implement KDGene with PyTorch and have made our source code available on GitHub.\footnote{\label{ft:code}https://github.com/sienna-wxy/KDGene} In our experiments, we carried out extensive grid search, over the following ranges of hyperparameter values: batch size in \{128, 256, 512, 1024\}, learning rate in \{0.01, 0.03, 0.05, 0.1\}, regularization coefficient in \{0.001, 0.01, 0.05, 0.1, 0.2, 0.5\}, the entity dimension in \{1000, 1500, 2000, 2500\} and the relation dimension in \{500, 1000, 1500, 2000\}. The optimal parameters used in KDGene can be seen on Github.\textsuperscript{\ref{ft:code}} Adagrad algorithm \cite{Adagrad} is adopted to optimize all trainable parameters. For all the baselines, we follow the best hyperparameters they provide and obtain disease gene prediction results on the DisGeNet dataset.

\subsection{Results and Analysis}
Table \ref{tab:result} reports the evaluation results of disease gene prediction on the DisGeNet dataset. It can be seen that KDgene outperforms all the baselines consistently and significantly. Specifically, compared with DGP baselines, KDGene realizes an average improvement of 16.59\% and over 25\% improvement on HR metric in particular. Compared with KGC baselines, in terms of HR@1, HR@3, HR@10, HR@50, MAP@1, MAP@3, MAP@10, MAP@50, the performance gains achieved by our model are 17.24\%, 21.44\%, 31.15\%, 20.35\%, 17.23\%, 21.12\%, 23.52\%, 23.59\%, with an average improvement of 21.96\%. These results illustrate the effectiveness of KDGene for disease gene prediction.

Our results also suggest that the success of KDGene does not mean that KGC-based models are the best in all DGP models (e.g., the MAP metric of GLIM\_DG is better than all existing KGC methods), but they can still outperform most DPG models. It's very promising for KGC-based methods to become the best in terms of the HR metric, showing a powerful recall ability, which indicates that combining complex relations in biological knowledge is beneficial for predicting candidate genes comprehensively.

Existing DGP models combined with KGE \cite{DG-KGE2} usually adopt conventional KGC methods. Our experiments confirm that existing KGC-based models are effective but not necessarily optimal. Among these typical KGC models, methods based on tensor decomposition perform better. Adopting tensor decomposition, we further introduce an interaction module based on the gating mechanism. It is worth noting that the result of CP-N3 can be regarded as an ablation study. Compared with CP-N3, KDGene realizes an increase by 39.86\%, 47.35\%, 52.12\%, 31.81\%, 39.85\%, 46.30\%, 47.52\%, and 45.39\% on HR@1, HR@3, HR@10, HR@50, MAP@1, MAP@3, MAP@10, and MAP@50, respectively, with an anveage improvement of 43.77\%. The impressive improvement compared with CP-N3 demonstrates the significance of our proposed interaction module, which enables KDGene with more precise representations of entities and relations to predict disease genes.

\subsection{Comparison of Different KGs}
\begin{figure}[h!] 
	\centering
	\includegraphics[scale=0.085]{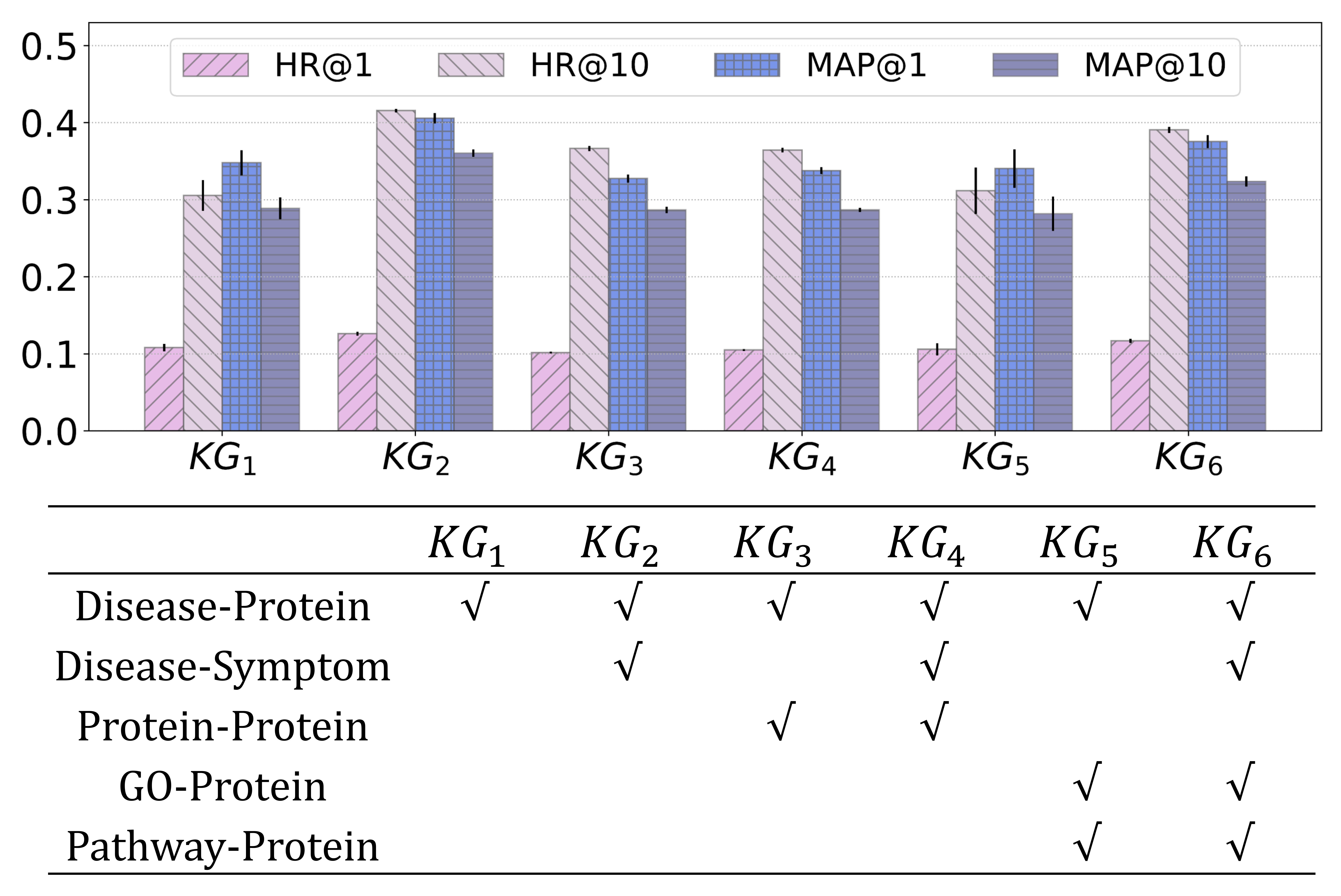}
	\caption{ Results of KDGene with different biological KGs. We use five relations associated with disease and gene to evaluate the effect of six combinations on KDGene performance. Except for the different KGs used, other experimental settings are kept the same. }
	\label{fig:KGs}
\end{figure}

To evaluate the impact of KGs composed of different relations on KDGene, we use different combinations of relations to construct biological KGs from $KG_1$ to $KG_6$ and evaluate the performance of KDGene. The results are shown in Figure \ref{fig:KGs}. $KG_1$ consists only of the disease-gene facts in the training set, with no external relations introduced. Based on $KG_1$, $KG_2$ and $KG_3$ introduce the facts about Disease-Symptom and Protein-Protein Interaction, respectively. $KG_2$ achieves the best performance, indicating that disease-symptom associations are beneficial for candidate gene prediction, while PPI has little effect. $KG_4$, which jointly introduces the two relations, fails to achieve the cumulative effect. Referring \cite{DG-KGE3}, $KG_5$ introduces GO and Pathway associations of genes. $KG_6$ reintroduces the disease-symptom relation based on $KG_5$. From the results, the former gets no obvious improvement, and the latter has improved, but still has not reached the performance of $KG_2$. The possible reason for the performance degradation of $KG_3$ and $KG_5$ that further introduce Protein-Protein or GO/Pathway-Protein associations is that the relations about proteins add more similar entities into KG, which could be noise.

\subsection{Comparison of Different PPI Score}
\begin{figure}[h!] 
	\centering
	\includegraphics[scale=0.07]{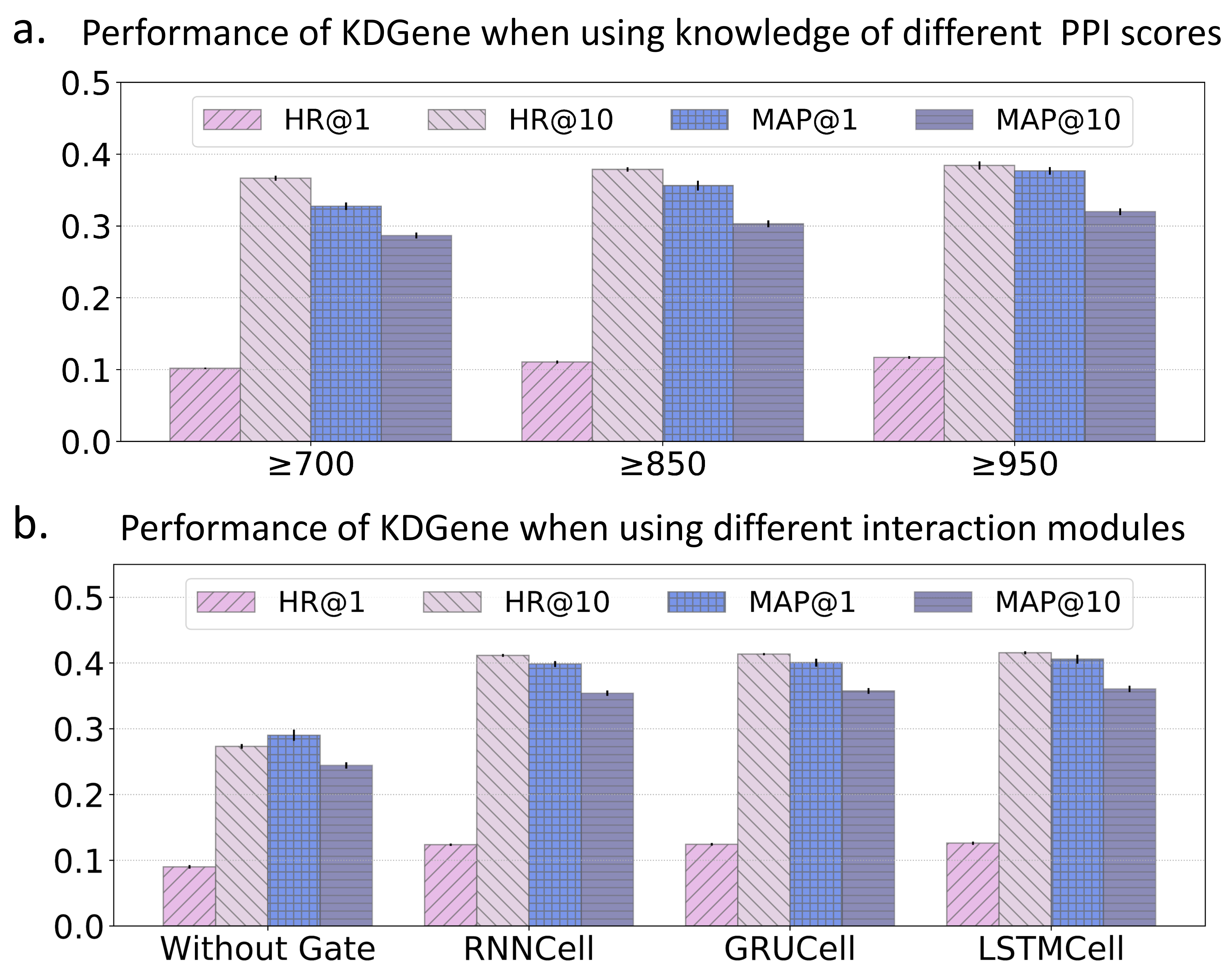}
	\caption{ \textbf{a.} Results of KDGene with facts in different confidence levels of PPI. \textbf{b.} Results of KDGene with different interaction modules.}
	\label{fig:PPI_Gates}
\end{figure}

To analyze the impact of knowledge with different confidence levels on KDGene, we consider scores in Protein-Protein Interaction facts. This score is often higher than the individual sub-scores, expressing increased confidence when an association is supported by several types of evidence \cite{STRING}. We select three grades of scores for evaluation, that is, the interaction scores $\ge700$, $\ge850$, and $\ge950$, respectively. For a fair comparison, all three evaluations are performed on the KG with the same disease-gene facts, with no differences other than the PPI facts. In Figure \ref{fig:PPI_Gates}a, as the score threshold increases, the performance of KDGene gradually improves, which indicates the introduction of reliable biological knowledge into the KG is more beneficial for KDGene to learn the representations of entities and relations. 

\subsection{Comparison of Different interaction modules}
To evaluate the impact of different interaction modules on the performance of KDGene, we conduct experiments with similar structures such as RNNCell and GRUCell, and the results are shown in Figure \ref{fig:PPI_Gates}b. Among the three gating mechanisms, the relation embedding is used as the input, and the head entity embedding as the hidden layer. The results of introducing different interaction structures are all better than the model without the gating mechanism (here we compare with CP-N3). The results of introducing different interaction structures are all better than the model without the gating mechanism (here we compare with CP-N3), illustrating the significance of the interaction module for tensor decomposition models. Among the three, the result of LSTMCell is slightly better than the remaining two. The possible reason is that the setting of the forget gate makes it more parameters to learn more details.

\begin{figure}[h!] 
	\centering
	\includegraphics[scale=0.35]{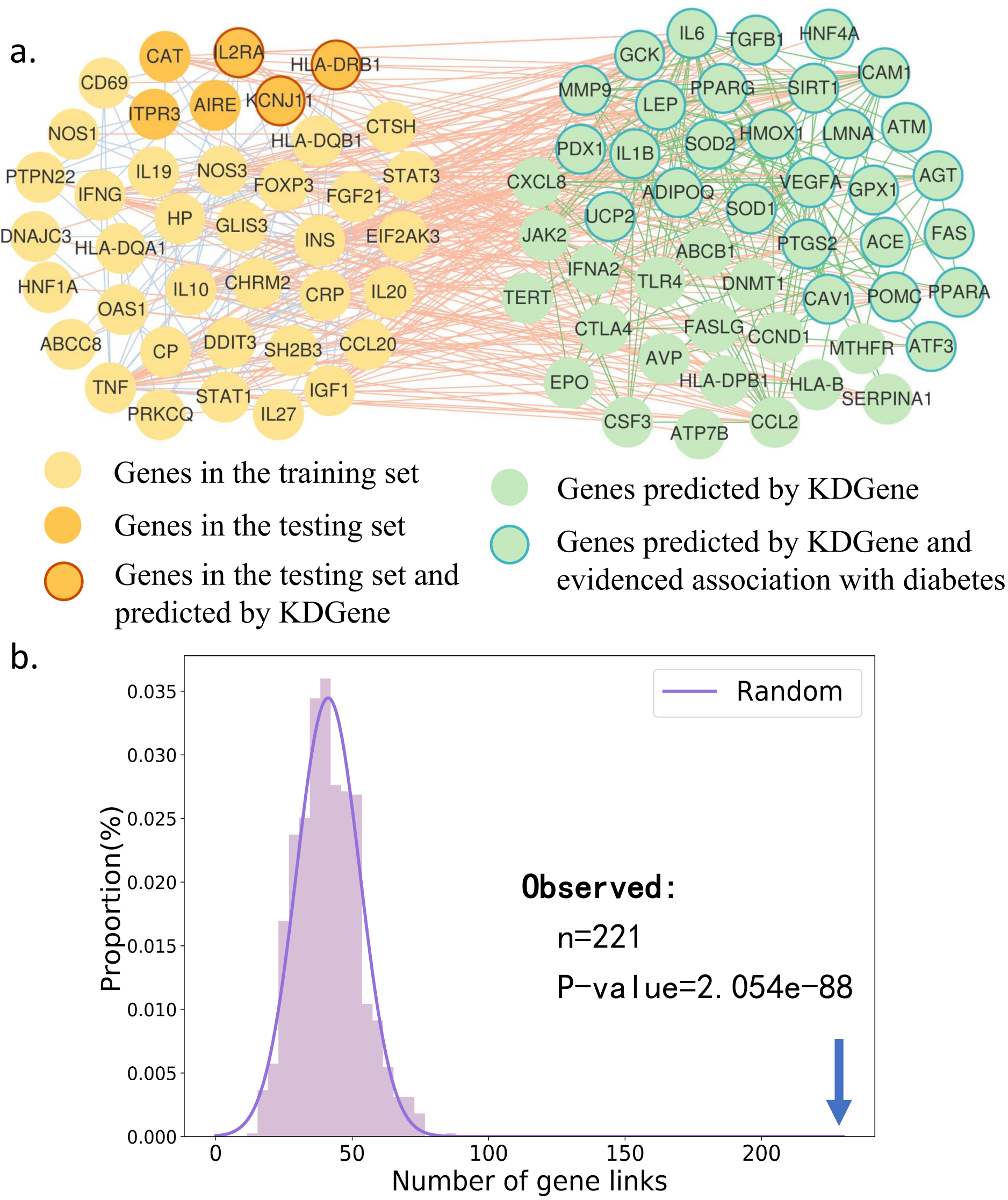}
	\caption{ PPI Network of the known and candidate genes of Diabetes Mellitus. The observed number of network links is significantly larger than the random control (P = 2.05E-88, binomial test).}
	\label{fig:case}
\end{figure}

\subsection{Case study}

We take Diabetes Mellitus as a case example to illustrate the high network closeness and functional relevance between genes in the training set and the candidate genes predicted by KDGene. For the disease, we keep all 42 genes in the training set and 6 genes in the testing set of the DisGeNet dataset and take out the top 50 candidate genes predicted by KDGene. The dense links (221 real links vs. 41.23 expected links, P = 2.05E-88, binomial test) that hold in the PPI network indicate that those two kinds of genes would tend to have closer interactions than expectation and rely on the same functional module in the PPI network.  The top ten gene prediction results of KDGene hit three diabetes mellitus-associated genes in the testing set. Meanwhile, we find that more than half of the predicted genes have corresponding literature evidence for association with diabetes. For example, interleukin-6 (IL-6, the top predicted gene), is not in the testing set, but \cite{IL6} indicates pro-inflammatory cytokines, such as interleukin-6 (IL-6), have been considered as key factors in type 1 diabetes mellitus (T1DM) and diabetic nephropathy. The results illustrate the accuracy and reliability of KDGene's prediction, which are promising to provide valuable references for further wet experiments.

\section{Conclusion}

In this paper, we first utilize the biological knowledge bases to create KGs and develop a scalable end-to-end knowledge graph completion model using an interactional tensor decomposition to identify disease-gene associations. KDGene introduces a gating mechanism-based interaction module between the embeddings of entities and relations to tensor decomposition, which can effectively enhance the information interaction in biological knowledge. Experimental results show that KDGene performs optimally among existing disease gene prediction methods. We also evaluate the impacts of KGs composed of knowledge with different relation types and degrees of confidence on KDGene's performance. Furthermore, the comprehensive biological analysis of the case of diabetes mellitus confirms KDGene's ability to identify new and accurate candidate genes. Future work will explore the capability of attention-based interaction modules in disease gene prediction. We will also extend this kind of module to other KGC models.



\bibliographystyle{named}
\bibliography{ijcai23}

\begin{thebibliography}{}

\bibitem[\protect\citeauthoryear{Alshahrani and Hoehndorf}{2018}]{ne}
Mona Alshahrani and Robert Hoehndorf.
\newblock Semantic disease gene embeddings (smudge): phenotype-based disease
  gene prioritization without phenotypes.
\newblock {\em Bioinformatics}, 34(17):i901--i907, 2018.

\bibitem[\protect\citeauthoryear{Ashley}{2016}]{precision_medicine}
Euan~A Ashley.
\newblock Towards precision medicine.
\newblock {\em Nature Reviews Genetics}, 17(9):507--522, 2016.

\bibitem[\protect\citeauthoryear{Bala{\v{z}}evi{\'c} \bgroup \em et al.\egroup
  }{2019}]{TuckER}
Ivana Bala{\v{z}}evi{\'c}, Carl Allen, and Timothy~M Hospedales.
\newblock Tucker: Tensor factorization for knowledge graph completion.
\newblock {\em arXiv preprint arXiv:1901.09590}, 2019.

\bibitem[\protect\citeauthoryear{Bordes \bgroup \em et al.\egroup
  }{2013}]{TransE}
Antoine Bordes, Nicolas Usunier, Alberto Garcia-Duran, Jason Weston, and Oksana
  Yakhnenko.
\newblock Translating embeddings for modeling multi-relational data.
\newblock {\em Advances in neural information processing systems}, 26, 2013.

\bibitem[\protect\citeauthoryear{Calvo \bgroup \em et al.\egroup
  }{2007}]{labor}
Borja Calvo, N{\'u}ria L{\'o}pez-Bigas, Simon~J Furney, Pedro Larra{\~n}aga,
  and Jose~A Lozano.
\newblock A partially supervised classification approach to dominant and
  recessive human disease gene prediction.
\newblock {\em Computer methods and programs in biomedicine}, 85(3):229--237,
  2007.

\bibitem[\protect\citeauthoryear{Care \bgroup \em et al.\egroup
  }{2009}]{clustering}
MA~Care, JR~Bradford, CJ~Needham, AJ~Bulpitt, and DR~Westhead.
\newblock Combining the interactome and deleterious snp predictions to improve
  disease gene identification.
\newblock {\em Human mutation}, 30(3):485--492, 2009.

\bibitem[\protect\citeauthoryear{Choi and Lee}{2019}]{DG-KGE4}
Wonjun Choi and Hyunju Lee.
\newblock Inference of biomedical relations among chemicals, genes, diseases,
  and symptoms using knowledge representation learning.
\newblock {\em IEEE Access}, 7:179373--179384, 2019.

\bibitem[\protect\citeauthoryear{Choi and Lee}{2021}]{DG-KGE1}
Wonjun Choi and Hyunju Lee.
\newblock Identifying disease-gene associations using a convolutional neural
  network-based model by embedding a biological knowledge graph with entity
  descriptions.
\newblock {\em Plos one}, 16(10):e0258626, 2021.

\bibitem[\protect\citeauthoryear{Dettmers \bgroup \em et al.\egroup
  }{2018}]{ConvE}
Tim Dettmers, Pasquale Minervini, Pontus Stenetorp, and Sebastian Riedel.
\newblock Convolutional 2d knowledge graph embeddings.
\newblock In {\em Proceedings of the AAAI conference on artificial
  intelligence}, volume~32, 2018.

\bibitem[\protect\citeauthoryear{Duchi \bgroup \em et al.\egroup
  }{2011}]{Adagrad}
John Duchi, Elad Hazan, and Yoram Singer.
\newblock Adaptive subgradient methods for online learning and stochastic
  optimization.
\newblock {\em Journal of Machine Learning Research}, 12:2121--2159, 2011.

\bibitem[\protect\citeauthoryear{Erten \bgroup \em et al.\egroup }{2011}]{DADA}
Sinan Erten, Gurkan Bebek, Rob~M Ewing, and Mehmet Koyut{\"u}rk.
\newblock Da da: degree-aware algorithms for network-based disease gene
  prioritization.
\newblock {\em BioData mining}, 4(1):1--20, 2011.

\bibitem[\protect\citeauthoryear{Gers \bgroup \em et al.\egroup }{2000}]{LSTM2}
Felix~A Gers, J{\"u}rgen Schmidhuber, and Fred Cummins.
\newblock Learning to forget: Continual prediction with lstm.
\newblock {\em Neural computation}, 12(10):2451--2471, 2000.

\bibitem[\protect\citeauthoryear{GO}{2021}]{GO}
The gene ontology resource: enriching a gold mine.
\newblock {\em Nucleic acids research}, 49(D1):D325--D334, 2021.

\bibitem[\protect\citeauthoryear{Grover and Leskovec}{2016}]{node2vec}
Aditya Grover and Jure Leskovec.
\newblock node2vec: Scalable feature learning for networks.
\newblock In {\em Proceedings of the 22nd ACM SIGKDD international conference
  on Knowledge discovery and data mining}, pages 855--864, 2016.

\bibitem[\protect\citeauthoryear{Guney and Oliva}{2012}]{GUILD}
Emre Guney and Baldo Oliva.
\newblock Exploiting protein-protein interaction networks for genome-wide
  disease-gene prioritization.
\newblock 2012.

\bibitem[\protect\citeauthoryear{Guo \bgroup \em et al.\egroup }{2019}]{bc02}
Yucheng Guo, Chen Bao, Dacheng Ma, Yubing Cao, Yanda Li, Zhen Xie, and Shao Li.
\newblock Network-based combinatorial crispr-cas9 screens identify synergistic
  modules in human cells.
\newblock {\em ACS synthetic biology}, 8(3):482--490, 2019.

\bibitem[\protect\citeauthoryear{Hirschhorn}{2009}]{gwsa}
Joel~N Hirschhorn.
\newblock Genomewide association studies—illuminating biologic pathways.
\newblock {\em New England journal of medicine}, 360(17):1699--1701, 2009.

\bibitem[\protect\citeauthoryear{Hitchcock}{1927}]{CPdecomposition}
Frank~L Hitchcock.
\newblock The expression of a tensor or a polyadic as a sum of products.
\newblock {\em Journal of Mathematics and Physics}, 6(1-4):164--189, 1927.

\bibitem[\protect\citeauthoryear{Hochreiter and Schmidhuber}{1997}]{LSTM}
Sepp Hochreiter and J{\"u}rgen Schmidhuber.
\newblock Long short-term memory.
\newblock {\em Neural computation}, 9(8):1735--1780, 1997.

\bibitem[\protect\citeauthoryear{Hou \bgroup \em et al.\egroup }{2022}]{GLIM}
Siyu Hou, Peng Zhang, Kuo Yang, Lan Wang, Changzheng Ma, Yanda Li, and Shao Li.
\newblock Decoding multilevel relationships with the human tissue-cell-molecule
  network.
\newblock {\em Briefings in Bioinformatics}, 2022.

\bibitem[\protect\citeauthoryear{Huang and Li}{2010}]{bc03}
Yezhou Huang and Shao Li.
\newblock Detection of characteristic sub pathway network for angiogenesis
  based on the comprehensive pathway network.
\newblock {\em BMC bioinformatics}, 11(1):1--9, 2010.

\bibitem[\protect\citeauthoryear{Jalilvand \bgroup \em et al.\egroup
  }{2019}]{jalilvand2019disease}
Ali Jalilvand, Behzad Akbari, Fatemeh~Zare Mirakabad, and Foad Ghaderi.
\newblock Disease gene prioritization using network topological analysis from a
  sequence based human functional linkage network.
\newblock {\em arXiv preprint arXiv:1904.06973}, 2019.

\bibitem[\protect\citeauthoryear{Kanehisa and Goto}{2000}]{KEGG}
Minoru Kanehisa and Susumu Goto.
\newblock Kegg: kyoto encyclopedia of genes and genomes.
\newblock {\em Nucleic acids research}, 28(1):27--30, 2000.

\bibitem[\protect\citeauthoryear{Lacroix \bgroup \em et al.\egroup
  }{2018}]{CP-N3}
Timoth{\'e}e Lacroix, Nicolas Usunier, and Guillaume Obozinski.
\newblock Canonical tensor decomposition for knowledge base completion.
\newblock In {\em International Conference on Machine Learning}, pages
  2863--2872. PMLR, 2018.

\bibitem[\protect\citeauthoryear{Lage \bgroup \em et al.\egroup }{2007}]{nf}
Kasper Lage, E~Olof Karlberg, Zenia~M St{\o}rling, P{\'a}ll~I Olason, Anders~G
  Pedersen, Olga Rigina, Anders~M Hinsby, Zeynep T{\"u}mer, Flemming Pociot,
  Niels Tommerup, et~al.
\newblock A human phenome-interactome network of protein complexes implicated
  in genetic disorders.
\newblock {\em Nature biotechnology}, 25(3):309--316, 2007.

\bibitem[\protect\citeauthoryear{Lander}{2011}]{HGP}
Eric~S Lander.
\newblock Initial impact of the sequencing of the human genome.
\newblock {\em Nature}, 470(7333):187--197, 2011.

\bibitem[\protect\citeauthoryear{Lei \bgroup \em et al.\egroup }{2021}]{bc01}
Yipin Lei, Shuya Li, Ziyi Liu, Fangping Wan, Tingzhong Tian, Shao Li, Dan Zhao,
  and Jianyang Zeng.
\newblock A deep-learning framework for multi-level peptide--protein
  interaction prediction.
\newblock {\em Nature Communications}, 12(1):5465, 2021.

\bibitem[\protect\citeauthoryear{Li and Patra}{2010}]{RWRH}
Yongjin Li and Jagdish~C Patra.
\newblock Genome-wide inferring gene--phenotype relationship by walking on the
  heterogeneous network.
\newblock {\em Bioinformatics}, 26(9):1219--1224, 2010.

\bibitem[\protect\citeauthoryear{Lin \bgroup \em et al.\egroup }{2020}]{KGNN}
Xuan Lin, Zhe Quan, Zhi-Jie Wang, Tengfei Ma, and Xiangxiang Zeng.
\newblock Kgnn: Knowledge graph neural network for drug-drug interaction
  prediction.
\newblock In {\em IJCAI}, volume 380, pages 2739--2745, 2020.

\bibitem[\protect\citeauthoryear{Luo \bgroup \em et al.\egroup
  }{2021}]{luo2021predicting}
Ping Luo, Bolin Chen, Bo~Liao, and Fang-Xiang Wu.
\newblock Predicting disease-associated genes: Computational methods,
  databases, and evaluations.
\newblock {\em Wiley Interdisciplinary Reviews: Data Mining and Knowledge
  Discovery}, 11(2):e1383, 2021.

\bibitem[\protect\citeauthoryear{Nunes \bgroup \em et al.\egroup
  }{2021}]{DG-KGE2}
Susana Nunes, Rita~T Sousa, and Catia Pesquita.
\newblock Predicting gene-disease associations with knowledge graph embeddings
  over multiple ontologies.
\newblock {\em arXiv preprint arXiv:2105.04944}, 2021.

\bibitem[\protect\citeauthoryear{Picart-Armada \bgroup \em et al.\egroup
  }{2019}]{propagation}
Sergio Picart-Armada, Steven~J Barrett, David~R Will{\'e}, Alexandre
  Perera-Lluna, Alex Gutteridge, and Benoit~H Dessailly.
\newblock Benchmarking network propagation methods for disease gene
  identification.
\newblock {\em PLoS computational biology}, 15(9):e1007276, 2019.

\bibitem[\protect\citeauthoryear{Pi{\~n}ero \bgroup \em et al.\egroup
  }{2016}]{DisGeNet}
Janet Pi{\~n}ero, {\`A}lex Bravo, N{\'u}ria Queralt-Rosinach, Alba
  Guti{\'e}rrez-Sacrist{\'a}n, Jordi Deu-Pons, Emilio Centeno, Javier
  Garc{\'\i}a-Garc{\'\i}a, Ferran Sanz, and Laura~I Furlong.
\newblock Disgenet: a comprehensive platform integrating information on human
  disease-associated genes and variants.
\newblock {\em Nucleic acids research}, page gkw943, 2016.

\bibitem[\protect\citeauthoryear{Rossi \bgroup \em et al.\egroup
  }{2021}]{review}
Andrea Rossi, Denilson Barbosa, Donatella Firmani, Antonio Matinata, and Paolo
  Merialdo.
\newblock Knowledge graph embedding for link prediction: A comparative
  analysis.
\newblock {\em ACM Transactions on Knowledge Discovery from Data (TKDD)},
  15(2):1--49, 2021.

\bibitem[\protect\citeauthoryear{Sun \bgroup \em et al.\egroup }{2019}]{Rotate}
Zhiqing Sun, Zhi-Hong Deng, Jian-Yun Nie, and Jian Tang.
\newblock Rotate: Knowledge graph embedding by relational rotation in complex
  space.
\newblock {\em arXiv preprint arXiv:1902.10197}, 2019.

\bibitem[\protect\citeauthoryear{Trouillon \bgroup \em et al.\egroup
  }{2016}]{ComplEx}
Th{\'e}o Trouillon, Johannes Welbl, Sebastian Riedel, {\'E}ric Gaussier, and
  Guillaume Bouchard.
\newblock Complex embeddings for simple link prediction.
\newblock In {\em International conference on machine learning}, pages
  2071--2080. PMLR, 2016.

\bibitem[\protect\citeauthoryear{Ururahy \bgroup \em et al.\egroup
  }{2015}]{IL6}
Marcela Abbott~Galv{\~a}o Ururahy, Karla Simone~Costa de~Souza, Yonara Monique
  da~Costa Oliveira, Melina~Bezerra Loureiro, Heglayne Pereira~Vital da~Silva,
  Francisco~Paulo Freire-Neto, Joao~Felipe Bezerra, Andre~Ducati Luchessi,
  Sonia~Quateli Doi, Rosario Dominguez~Crespo Hirata, et~al.
\newblock Association of polymorphisms in il6 gene promoter region with type 1
  diabetes and increased albumin-to-creatinine ratio.
\newblock {\em Diabetes/metabolism research and reviews}, 31(5):500--506, 2015.

\bibitem[\protect\citeauthoryear{Vanunu \bgroup \em et al.\egroup
  }{2010}]{PRINCE}
Oron Vanunu, Oded Magger, Eytan Ruppin, Tomer Shlomi, and Roded Sharan.
\newblock Associating genes and protein complexes with disease via network
  propagation.
\newblock {\em PLoS computational biology}, 6(1):e1000641, 2010.

\bibitem[\protect\citeauthoryear{Vilela \bgroup \em et al.\egroup
  }{2022}]{DG-KGE3}
Joana Vilela, Muhammad Asif, Ana~Rita Marques, Jo{\~a}o~Xavier Santos,
  C{\'e}lia Rasga, Astrid Vicente, and Hugo Martiniano.
\newblock Biomedical knowledge graph embeddings for personalized medicine:
  Predicting disease-gene associations.
\newblock {\em Expert Systems}, page e13181, 2022.

\bibitem[\protect\citeauthoryear{Von~Mering \bgroup \em et al.\egroup
  }{2005}]{STRING}
Christian Von~Mering, Lars~J Jensen, Berend Snel, Sean~D Hooper, Markus Krupp,
  Mathilde Foglierini, Nelly Jouffre, Martijn~A Huynen, and Peer Bork.
\newblock String: known and predicted protein--protein associations, integrated
  and transferred across organisms.
\newblock {\em Nucleic acids research}, 33(suppl\_1):D433--D437, 2005.

\bibitem[\protect\citeauthoryear{Wang \bgroup \em et al.\egroup
  }{2017}]{survey}
Quan Wang, Zhendong Mao, Bin Wang, and Li~Guo.
\newblock Knowledge graph embedding: A survey of approaches and applications.
\newblock {\em IEEE Transactions on Knowledge and Data Engineering},
  29(12):2724--2743, 2017.

\bibitem[\protect\citeauthoryear{Wu \bgroup \em et al.\egroup }{2008a}]{bc05}
Xuebing Wu, Rui Jiang, Michael~Q Zhang, and Shao Li.
\newblock Network-based global inference of human disease genes.
\newblock {\em Molecular systems biology}, 4(1):189, 2008.

\bibitem[\protect\citeauthoryear{Wu \bgroup \em et al.\egroup
  }{2008b}]{wu2008network}
Xuebing Wu, Rui Jiang, Michael~Q Zhang, and Shao Li.
\newblock Network-based global inference of human disease genes.
\newblock {\em Molecular systems biology}, 4(1):189, 2008.

\bibitem[\protect\citeauthoryear{Wu \bgroup \em et al.\egroup }{2019a}]{SymMap}
Yang Wu, Feilong Zhang, Kuo Yang, Shuangsang Fang, Dechao Bu, Hui Li, Liang
  Sun, Hairuo Hu, Kuo Gao, Wei Wang, et~al.
\newblock Symmap: an integrative database of traditional chinese medicine
  enhanced by symptom mapping.
\newblock {\em Nucleic acids research}, 47(D1):D1110--D1117, 2019.

\bibitem[\protect\citeauthoryear{Wu \bgroup \em et al.\egroup }{2019b}]{Renet}
Ye~Wu, Ruibang Luo, Henry Leung, Hing-Fung Ting, and Tak-Wah Lam.
\newblock Renet: A deep learning approach for extracting gene-disease
  associations from literature.
\newblock In {\em International Conference on Research in Computational
  Molecular Biology}, pages 272--284. Springer, 2019.

\bibitem[\protect\citeauthoryear{Yang \bgroup \em et al.\egroup
  }{2014}]{DistMult}
Bishan Yang, Wen-tau Yih, Xiaodong He, Jianfeng Gao, and Li~Deng.
\newblock Embedding entities and relations for learning and inference in
  knowledge bases.
\newblock {\em arXiv preprint arXiv:1412.6575}, 2014.

\bibitem[\protect\citeauthoryear{Yang \bgroup \em et al.\egroup
  }{2018a}]{network}
Kuo Yang, Ning Wang, Guangming Liu, Ruyu Wang, Jian Yu, Runshun Zhang, Jianxin
  Chen, and Xuezhong Zhou.
\newblock Heterogeneous network embedding for identifying symptom candidate
  genes.
\newblock {\em Journal of the American Medical Informatics Association},
  25(11):1452--1459, 2018.

\bibitem[\protect\citeauthoryear{Yang \bgroup \em et al.\egroup
  }{2018b}]{HerGePred}
Kuo Yang, Ruyu Wang, Guangming Liu, Zixin Shu, Ning Wang, Runshun Zhang, Jian
  Yu, Jianxin Chen, Xiaodong Li, and Xuezhong Zhou.
\newblock Hergepred: heterogeneous network embedding representation for disease
  gene prediction.
\newblock {\em IEEE journal of biomedical and health informatics},
  23(4):1805--1815, 2018.

\bibitem[\protect\citeauthoryear{Yang \bgroup \em et al.\egroup
  }{2020}]{PDGNet}
Kuo Yang, Yi~Zheng, Kezhi Lu, Kai Chang, Ning Wang, Zixin Shu, Jian Yu, Baoyan
  Liu, Zhuye Gao, and Xuezhong Zhou.
\newblock Pdgnet: Predicting disease genes using a deep neural network with
  multi-view features.
\newblock {\em IEEE/ACM Transactions on Computational Biology and
  Bioinformatics}, 2020.

\bibitem[\protect\citeauthoryear{Yang \bgroup \em et al.\egroup
  }{2021}]{MapGene}
Kuo Yang, Kezhi Lu, Yang Wu, Jian Yu, Baoyan Liu, Yi~Zhao, Jianxin Chen, and
  Xuezhong Zhou.
\newblock A network-based machine-learning framework to identify both
  functional modules and disease genes.
\newblock {\em Human Genetics}, 140(6):897--913, 2021.

\bibitem[\protect\citeauthoryear{Yu \bgroup \em et al.\egroup
  }{2019}]{LSTMreview}
Yong Yu, Xiaosheng Si, Changhua Hu, and Jianxun Zhang.
\newblock A review of recurrent neural networks: Lstm cells and network
  architectures.
\newblock {\em Neural computation}, 31(7):1235--1270, 2019.

\bibitem[\protect\citeauthoryear{Zhang \bgroup \em et al.\egroup }{2020}]{bc04}
Yiding Zhang, Lyujie Chen, and Shao Li.
\newblock Cipher-sc: disease-gene association inference using graph convolution
  on a context-aware network with single-cell data.
\newblock {\em IEEE/ACM Transactions on Computational Biology and
  Bioinformatics}, 19(2):819--829, 2020.

\bibitem[\protect\citeauthoryear{Zhu \bgroup \em et al.\egroup
  }{2022}]{zhu2022multimodal}
Chaoyu Zhu, Zhihao Yang, Xiaoqiong Xia, Nan Li, Fan Zhong, and Lei Liu.
\newblock Multimodal reasoning based on knowledge graph embedding for specific
  diseases.
\newblock {\em Bioinformatics}, 38(8):2235--2245, 2022.

\end{thebibliography}

\end{document}